%% file: main.tex
\documentclass[10pt,twocolumn,letterpaper]{article}

\usepackage{cvpr}              

\input{preamble}

\definecolor{cvprblue}{rgb}{0.21,0.49,0.74}
\usepackage[pagebackref,breaklinks,colorlinks,citecolor=cvprblue]{hyperref}

\usepackage{graphicx}
\usepackage{amsmath}
\usepackage{amssymb}
\usepackage{booktabs}
\usepackage[T1]{fontenc}
\usepackage{multirow}
\usepackage{makecell}
\usepackage{amssymb}
\usepackage{bbm}
\usepackage{bbding}
\usepackage{colortbl}
\usepackage{threeparttable}
\usepackage{eufrak}
\usepackage{makecell}
\usepackage{pifont}

\usepackage[pagebackref,breaklinks,colorlinks]{hyperref}
\usepackage{stmaryrd}
\usepackage{amssymb}

\newcommand{\keypoint}[1]{\vspace{0.1cm}\noindent\textbf{#1}}
\usepackage{framed}

\usepackage[capitalize]{cleveref} 
\crefname{section}{Sec.}{Secs.}
\Crefname{section}{Section}{Sections}
\Crefname{table}{Table}{Tables}
\crefname{table}{Tab.}{Tabs.}

\def\paperID{100} 
\def\confName{CVPR}
\def\confYear{2024}

\title{DePT: Decoupled Prompt Tuning} 

\author{
Ji Zhang\textsuperscript{1}\thanks{Equal contribution.} \quad  \quad 
Shihan Wu\textsuperscript{1}{\textcolor{black!70}{$^{\ast}$}} \quad  \quad 
 Lianli Gao\textsuperscript{2} \quad  \quad 
 Heng Tao Shen\textsuperscript{3} \quad  \quad  Jingkuan Song\textsuperscript{3,1}\thanks{Corresponding author.} \\
 \textsuperscript{1}University of Electronic Science and Technology of China (UESTC)\\
 \textsuperscript{2}Shenzhen Institute for Advanced Study, UESTC \\
 \textsuperscript{3}Tongji University \\
 {\tt\small \textcolor{magenta}{\href{}{\{jizhang.jim,jingkuan.song\}@gmail.com}}}
}

\begin{document}
\maketitle

\begin{abstract}
This work breaks through the Base-New Tradeoff (BNT) dilemma in prompt tuning, i.e., the better the tuned model generalizes to the base (or target) task, the worse it generalizes to new tasks, and vice versa. 
Specifically, through an in-depth analysis of the learned features of the base and new tasks, 
we observe that the BNT stems from a channel bias issue -- the vast majority of feature channels are occupied by base-specific knowledge, leading to the collapse of task-shared knowledge important to new tasks.
To address this, we propose the \textbf{De}coupled \textbf{P}rompt \textbf{T}uning (\textbf{DePT}) framework, which decouples base-specific knowledge from feature channels into an isolated feature space during prompt tuning, so as to maximally preserve task-shared knowledge in the original feature space for achieving better zero-shot generalization on new tasks.
Importantly, 
our DePT is orthogonal to existing prompt tuning approaches, and can enhance them with negligible additional computational cost.
Extensive experiments on several datasets show the flexibility and effectiveness of DePT.
Code is available at \url{https://github.com/Koorye/DePT}.
\end{abstract}

\section{Introduction}
\label{s1}

Recent years have witnessed remarkable progress in large vision-language pre-trained models (VLPMs).
One of the striking successes has been achieved by the contrastive language-image pretraining (CLIP) \cite{radford2021clip} model, which formulates the learning objective as a contrastive loss to establish alignment between images and their textual descriptions in a common feature space.
Despite the ability to capture open-set visual concepts, the zero-shot generalization performance of VLPMs is greatly reduced when there is a severe \textit{category shift}, \textit{distribution shift}, or \textit{domain shift} between upstream training data and downstream tasks.

Inspired by the success of prompt engineering in NLP, \textit{Prompt Tuning} (or, \textit{Context Optimization} \cite{zhou2022learning}) has emerged as a parameter-efficient learning paradigm to adapt powerful VLPMs to downstream tasks, by optimizing a task-specific prompt (i.e., a set of trainable vectors) with a handful of training data from the base (target) task while keeping the weights of VLPMs frozen.
Although the advantages are remarkable, existing prompt tuning methods usually fail to escape the Base-New Tradeoff (BNT)  dilemma, i.e., the better the tuned/adapted model generalizes to the base task, the worse it generalizes to new tasks (with unseen classes), and vice versa. 
Numerous efforts \cite{zhou2022conditional,yao2023visual,zhu2022prompt} have been devoted in recent years to alleviate the performance degradation of tuned models on new tasks by developing anti-overfitting strategies in the process of prompt tuning. 
Nevertheless, the BNT problem is still far from being resolved and its underlying mechanisms are poorly understood.

\begin{figure}[t]
\setlength{\abovecaptionskip}{0.2cm}  
\setlength{\belowcaptionskip}{-0.5cm} 
\centering
\includegraphics[width=0.85\linewidth]{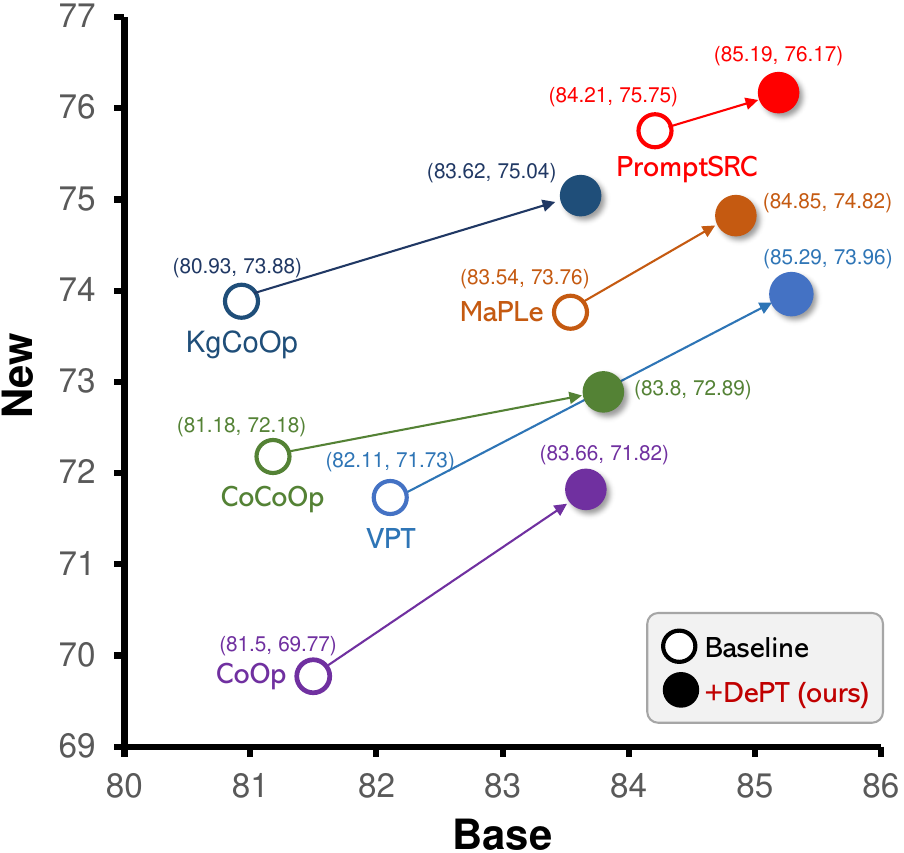} 
\caption{
Classification ACCs of six prompt tuning methods w/ or w/o our DePT framework on \textbf{Base} (or {seen}) and \textbf{New} ({or unseen}) tasks, averaged over 11 datasets in Table \ref{table—B2N}.
} 
\label{f1}
\end{figure}

\begin{figure*}[h]
\setlength{\abovecaptionskip}{0.2cm}  
\setlength{\belowcaptionskip}{-0.4cm} 
\centering
\includegraphics[width=0.9\linewidth]{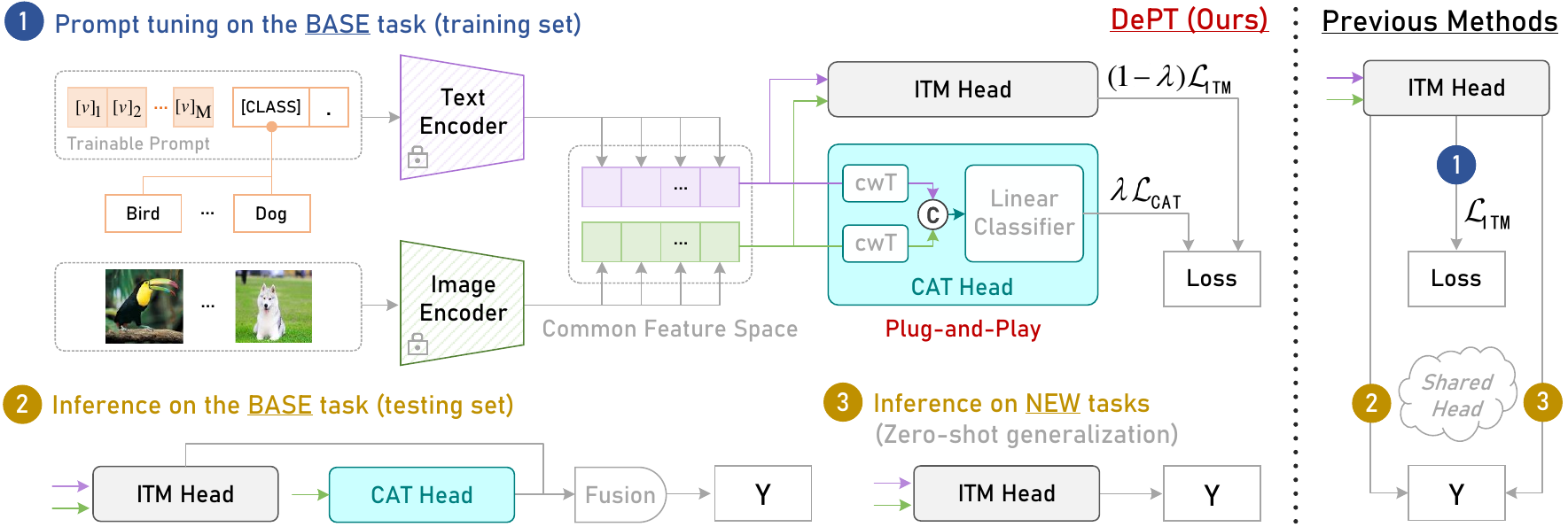} 
\caption{Illustration of our DePT framework (in a CoOp \cite{zhou2022learning} style). 
Unlike previous methods (\textit{right}) that use the same Image Text Matching (\textbf{ITM}) head for training/inference on the base task and zero-shot generalization on new tasks, our {DePT} (\textit{left}) employs a Channel Adjusted Transfer (\textbf{CAT}) head to capture \textit{base-specific} knowledge in an isolated feature space,
so as to maximally preserve \textit{task-shared} knowledge in the original feature space for improving zero-shot generalization on new tasks.
At inference, we further boost the performance on the base task by simply fusing {base-specific} and {task-shard} knowledge obtained by the two heads.  \textcolor{black}{${\copyright}$} denotes the concatenation operation.} 
\label{pip}
\end{figure*}

In this work, we bridge the gap by proposing \textbf{De}coupled \textbf{P}rompt \textbf{T}uning (\textbf{DePT}), a first framework tackling the BNT problem in prompt tuning from a feature decoupling perspective.
Specifically, through an in-depth analysis of the feature channels of base and new tasks learned by the standard Image Text Matching (ITM) head, 
we discern that the BNT stems from a \textit{channel bias} issue: the vast majority of feature channels are occupied by \textit{base-specific} knowledge (i.e., task-specific knowledge of the base task), resulting in the collapse of \textit{task-shared} knowledge important to new tasks (Section \ref{sec_mtv}).
Inspired by this, the direct strategy to tackle the BNT problem is to decouple base-specific knowledge and task-shared knowledge in feature channels during prompt tuning.
To accomplish this, we introduce a Channel Adjusted Transfer (CAT) head to encourage the mining of base-specific knowledge from feature channels in an isolated feature space, 
thereby facilitating the preservation of task-shared knowledge in the original feature space and enhancing zero-shot generalization performance on new tasks (Section \ref{sect_dept}).
Furthermore, by simply fusing base-specific knowledge and task-shard knowledge in the two feature spaces at inference, we boost the performance on the base task remarkably (Section \ref{ablation}).

\keypoint{Flexibility and Effectiveness.} 
Our DePT framework is orthogonal to existing prompt tuning methods, hence it can be flexibly used to overcome the BNT problem for them. 
We evaluate our DePT using a broad spectrum of baseline methods, including the \textit{visual} prompt tuning method VPT \cite{jia2022vpt}, \textit{textual} prompt tuning methods CoOp \cite{zhou2022learning}, CoCoOp \cite{zhou2022conditional} and KgCoOp \cite{yao2023visual}, and \textit{multi-model} prompt tuning methods MaPLe \cite{khattak2023maple}, PromptSRC \cite{Khattak_2023_ICCV}. 
Experimental results on 11 diverse datasets show that DePT consistently improves the performance of those methods, regardless of whether there is a \textit{category shift} \textit{distribution shift} or \textit{domain shift} between base and new tasks, demonstrating the strong flexibility and effectiveness of DePT (Section \ref{expres}).
Notably, DePT enhances the six baselines without performance tradeoffs on base and new tasks -- DePT achieves absolute gains of \textbf{1.31}\%$\thicksim$\textbf{3.17}\% (resp. \textbf{0.71}\%$\thicksim$\textbf{2.23}\%) on base  (resp. new) tasks, averaged on the 11 datasets (Figure \ref{f1}).


\keypoint{Contributions.} 
Our main contributions are threefold.
\textbf{1)} We provide an insightful analysis of the BNT problem in prompt tuning, revealing for the first time that the BNT stems from the channel bias issue.
\textbf{2)} We propose the DePT framework to tackle the BNT problem from a feature decoupling perspective, and DePT is orthogonal to existing prompt tuning methods.
\textbf{3)} We perform experiments on 11 diverse datasets and show that DePT consistently enhances the performance of a broad spectrum of baseline methods.\footnote{Our proposed DePT can also be used as a plugin to improve existing {\textit{adapter}} tuning methods, as proven in \underline{\textbf{Sup.Mat. (E)}}.}

\begin{figure*}[t]
\setlength{\abovecaptionskip}{0.1cm}  
\setlength{\belowcaptionskip}{-0.3cm} 
\centering
\includegraphics[width=1\linewidth]{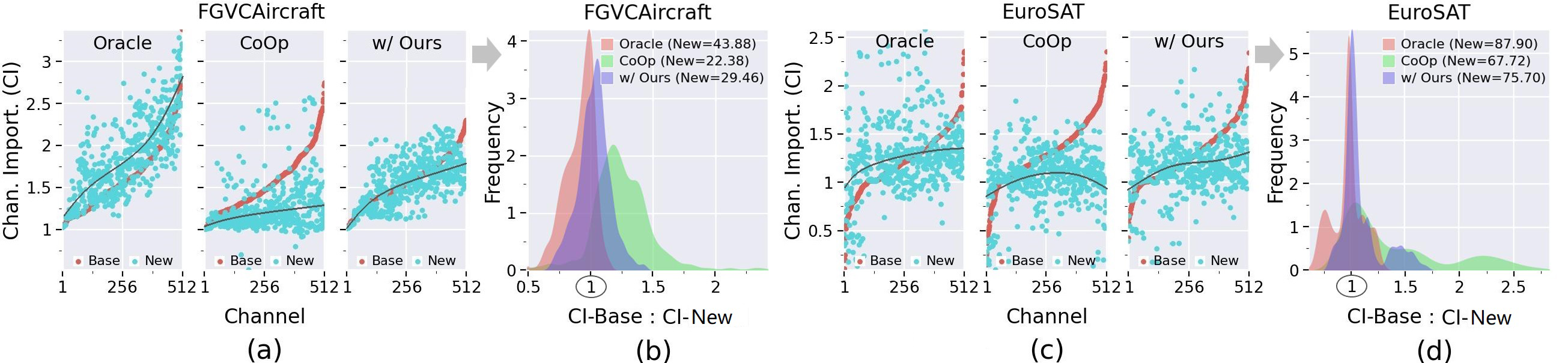} 
\caption{Channel Importance (\textbf{CI}) distributions of base and new tasks learned by the Oracle model and CoOp \cite{zhou2022learning} w/ or w/o our DePT on the datasets FGVCAircraft \cite{maji2013fine} and  EuroSAT \cite{helber2019eurosat}.
In {(\textbf{a})(\textbf{c})}, 
the indexes of channels in the \textit{x}-axis are reordered based on the CI of the base task, 
a blue/red point indicates a channel. In {(\textbf{b})(\textbf{d})}, the frequency distributions of CI-Base : CI-New are presented, where CI-Base and CI-New are the CI of the base and new tasks, respectively; 
“H” denotes the harmonic mean \cite{zhou2022conditional} of base-task and new-task accuracies.
} 
\label{f2}
\end{figure*}

\section{Methodology}
In this section, we initially offer an insightful examination of the BNT problem in prompt tuning, followed by a detailed exposition of our proposed DePT framework.

\subsection{Preliminaries}
\label{prelim}
\keypoint{Contrastive Language-Image Pre-training (CLIP)  \cite{wang2022clip}.}
CLIP targets learning an alignment between image and text features produced by an image encoder and a text encoder, respectively.
After seeing 400 million image-text association pairs and performing a contrastive learning paradigm in a common feature space, CLIP captures diverse open-set visual concepts that can readily be generalized to downstream applications.
For example, we can achieve zero-shot classification by formulating the classification task as an image-text matching problem.
Specifically, we first craft a prompt (e.g., “\texttt{a photo of a}”) to obtain the text features of all inner-task classes by feeding the class-extended prompt (e.g., “\texttt{a photo of a [CLASS]}”) to the text encoder.
Then, we use the image encoder to obtain the image feature of an input example, and predict the class of the example by comparing the cosine distances between the image feature and the text features of classes.
\keypoint{Prompt Tuning with the Image-Text Matching Head.} 
Instead of using a hand-crafted prompt (e.g., “\texttt{a photo of a}”), prompt tuning aims to learn a \textit{task-specific} prompt using a handful of training data from the base (or target) task.
Let $[\boldsymbol{v}]_1[\boldsymbol{v}]_2...[\boldsymbol{v}]_l$ denote $l$ trainable vectors, we forward the class-extended prompt $\boldsymbol{{c}}_i=[\boldsymbol{v}]_1[\boldsymbol{v}]_2...[\boldsymbol{v}]_l[\texttt{CLASS}]$ to the text encoder $g(\cdot)$ to obtain the text feature of the $i$-th class: $g(\boldsymbol{{c}}_i)$.
Let $\boldsymbol{f}$ denote the image feature of an example $\boldsymbol{x}$ obtained by the image encoder, the task-specific prompt can be optimized using a parameter-free Image-Text Matching (ITM) head, which formulates the learning objective as:
\begin{equation}
    \mathcal{L}_{\mathtt{ITM}}=-\sum_i \boldsymbol{y}_i \log \mathcal{P}_{\mathtt{ITM}}(\boldsymbol{{c}}_i| \boldsymbol{x}),
\end{equation}
where $\boldsymbol{y}$ is the one-hot label, 
\begin{equation}
    \mathcal{P}_{\mathtt{ITM}}(\boldsymbol{{c}}_i | \boldsymbol{x})=\frac{\exp(<g(\boldsymbol{{c}}_i), \boldsymbol{f}>/\tau)}{\sum_{i=1}^{M}\exp(<g(\boldsymbol{{c}}_i), \boldsymbol{f}>/\tau)},
    \label{inf}
\end{equation}
 $<\cdot>$ denotes cosine similarity, $M$ is the number of classes, and $\tau$ is the temperature learned by CLIP.
During training, the gradients calculated in the ITM head can be back-propagated all the way through the text encoder $g({\cdot})$ to optimize the trainable vectors in the prompt.

\subsection{A Closer Look at the BNT Problem}
\label{sec_mtv}
Due to the BNT problem, adapting a pretrained model to the base task $\mathcal{T}_{\mathrm{base}}$ will decrease the generalization of the model on the new task $\mathcal{T}_{\mathrm{new}}$, and {vice versa}.
In this part, we provide an insightful view to analyze the BNT problem. 

\keypoint{Deriving an Oracle Model on $\mathcal{T}_{\mathrm{base}}$ and $\mathcal{T}_{\mathrm{new}}$.}  
We start the investigation of the BNT problem by deriving an \textit{oracle} model on  $\mathcal{T}_{\mathrm{base}}$ and $\mathcal{T}_{\mathrm{new}}$.
Specifically, we adapt the pretrained model to both $\mathcal{T}_{\mathrm{base}}$ and $\mathcal{T}_{\mathrm{new}}$ by jointly training the model on the data of the two tasks during prompt tuning. The derived oracle model thus can be seen as an approximation of a \textit{BNT-free} model, since it avoids overfitting to either $\mathcal{T}_{\mathrm{base}}$ or $\mathcal{T}_{\mathrm{new}}$. 
Here, we use the word “oracle”, because the model is derived by leveraging data from the new task, which is not accessible in prompt tuning.

\keypoint{Calculating Channel Importance for $\mathcal{T}_{\mathrm{base}}$ 
and $\mathcal{T}_{\mathrm{new}}$.} 
Denote ${\boldsymbol{f}}_j$ and $\boldsymbol{{e}}_* \in \{\boldsymbol{{e}}_i=g({\boldsymbol{{c}}_i})\}_{i=1}^{M}$ the $d$-dimensional image and text features of the example $\boldsymbol{x}_j$ in the learned feature space, respectively.
We calculate the {Channel Importance} (\textbf{CI}) of the $r$-th ($r=1,...,d$) feature channel for each task of $\mathcal{T}_{\mathrm{base}}$ and $\mathcal{T}_{\mathrm{new}}$ as follows:
\begin{equation}
    \boldsymbol{\mathrm{CI}}^{(r)} = \frac{1}{N} \sum_{j=1}^N \frac{\mathrm{{ReLU}}({\bar{\boldsymbol{{e}}}_{*}^{(r)}}{\bar{\boldsymbol{f}}}_{j}^{(r)})}{1/M \sum_{i=1}^{M}{\mathrm{ReLU}}({\bar{\boldsymbol{{e}}}_{i}^{(r)}}{\bar{\boldsymbol{f}}}_{j}^{(r)})},
    \label{ci}
\end{equation}
where $\bar{{\cdot}}={{\cdot}}/{||{\cdot}||_2}$, $N$ is the number of examples in the task. $\mathrm{ReLU}$ \cite{agarap2018relu} is used to avoid the denominator being equal to 0.
The derived Eq. (\ref{ci})  has an intuitive explanation: 
a feature channel is of greater importance if it can better distinguish the classes in the task, 
i.e., the image features are close to the ground-truth text features and far away from the text features of other classes at this channel. 

\keypoint{Analysis.}
What are the differences between the derived oracle model and the model learned through the standard prompt tuning paradigm w.r.t. the calculated CI distributions of $\mathcal{T}_{\mathrm{base}}$ and $\mathcal{T}_{\mathrm{new}}$?
To answer this question, we take CoOp \cite{zhou2022learning} as the baseline method, and plot the CI distributions of the testing data of $\mathcal{T}_{\mathrm{base}}$ and $\mathcal{T}_{\mathrm{new}}$ for CoOp and the oracle model on the datasets FGVCAircraft \cite{maji2013fine} and EuroSAT \cite{helber2019eurosat} in Figure \ref{f2} (see \underline{\textbf{Sup. Mat.(A)}} for details).   
As observed in the figure, the CI distributions of base and new tasks obtained by the oracle model show greater consistency compared to that obtained by CoOp. 
Concretely, from the results of CoOp in \textbf{(a)(c)}, the achieved CI values of new tasks are significantly lower than that of base tasks at the vast majority of feature channels, which is further confirmed in \textbf{(b)(d)}, where the computed values of “CI-Base : CI-New” are larger than (resp. close to) \textbf{{1.0}} in most cases for CoOp (resp. the oracle model).
In \textbf{(b)(d)}, we present the classification accuracies of CoOp and the oracle model on new tasks, where the oracle model outperforms CoOp by large margins, suggesting that most feature channels produced by the oracle model contain \textit{task-shared} knowledge that is valuable for the generalization of new tasks. 
In a nutshell, after prompt tuning, the vast majority of learned feature channels are occupied by \textit{base-specific} knowledge, resulting in the collapse (or catastrophic forgetting) of {task-shared} knowledge important to new tasks -- we refer to this as a \textit{channel bias} issue in this work.
Inspired by the above observations, we raise the following question:
\begin{framed}
\textit{Can we simultaneously preserve base-specific and task-shared knowledge in feature channels to overcome the BNT problem in prompt tuning?}
\end{framed}

\subsection{Decoupled Prompt Tuning}
\label{sect_dept}
In this work, we answer the above question by proposing Decoupled Prompt Tuning (DePT), a first framework overcoming the BNT problem in prompt tuning from a feature decoupling perspective.
An overview of the DePT framework is presented in Figure \ref{pip}.

\keypoint{A Plug-and-play Channel Adjusted Transfer Head.}
Due to the channel bias issue, striving for base-specific knowledge during prompt tuning will inevitably trigger the catastrophic forgetting of task-shared knowledge in the learned feature channels.
To address this, DePT employs a Channel Adjusted Transfer (CAT) head to decouple base-specific knowledge from feature channels into an isolated feature space, so as to maximally preserve task-shared knowledge in the original feature space.
Denote $\mathcal{S}_{\mathtt{img}}=\{\boldsymbol{f}_j\}_{j=1}^{J}$ and $\mathcal{S}_{\mathtt{text}}=\{\boldsymbol{e}_j\}_{j=1}^{J}$\footnote{Here, $\boldsymbol{e}_i$ may equal to $\boldsymbol{e}_j$ for $i \neq j$, we ignore it for simplification.} the sets of image and text features for a batch of training examples respectively, and $\boldsymbol{f}_j$, $\boldsymbol{e}_j \in \mathbb{R}^d$.
First, the CAT head leverages a {channel-wise Transformation} (cwT) layer to transform both $\mathcal{S}_{\mathtt{img}}$ and $\mathcal{S}_{\mathtt{text}}$ to a new feature space. 
Formally, $\mathcal{S}_{\mathtt{img}}^{\prime}=\{\boldsymbol{f}_j^{\prime}\}_{j=1}^{J}$, and
\begin{equation} 
    \boldsymbol{f}_j^{\prime} = \boldsymbol{\gamma} \odot \boldsymbol{f}_j + \boldsymbol{\beta}, \quad j= 1,...,J,
\end{equation}
where $\boldsymbol{\gamma}$, $\boldsymbol{\beta} \in \mathbb{R}^d$ are trainable scaling and shift vectors.
Denote $\mathcal{S}_{\mathtt{text}}^{\prime}=\{\boldsymbol{e}_j^{\prime}\}_{j=1}^{J}$ similar to $\mathcal{S}_{\mathtt{img}}^{\prime}=\{\boldsymbol{f}_j^{\prime}\}_{j=1}^{J}$. 
Next, a linear classifier takes $\mathcal{S}_{\cup}$ and $\mathcal{Y}_{\cup}$ as input to encourage the mining of base-specific knowledge in the isolated feature space, where $\mathcal{S}_{\cup}=\mathcal{S}_{\mathtt{img}}^{\prime} \cup \mathcal{S}_{\mathtt{text}}^{\prime}=\{\boldsymbol{s}_j\}_{j=1}^{2J}$ and $\mathcal{Y}_{\cup}=\{\boldsymbol{y}_j\}_{j=1}^{2J}$, $\boldsymbol{y}_j \in \mathbb{R}^{M}$ is the one-hot label for $\boldsymbol{s}_j$, and $M$ is the number of classes of the task.
For each pair of ($\boldsymbol{s}$, $\boldsymbol{y}$), the CAT head minimizes the following cross-entropy loss:
\begin{equation}
    \mathcal{L}_{\mathtt{CAT}}=-\sum_i \boldsymbol{y}_i \log \mathcal{P}_{\mathtt{CAT}}({\boldsymbol{c}}_i| \boldsymbol{x}),
\end{equation}
where 
\begin{equation}
    \mathcal{P}_{\mathtt{CAT}}(\boldsymbol{c}_i| \boldsymbol{x})=\sigma(\boldsymbol{W} \!\! \cdot \! \boldsymbol{s})[i],
    \label{ddd}
\end{equation}
$\boldsymbol{W} \in \mathbb{R}^{M \times d}$ denotes the projection matrix for classification, $\sigma$ denotes the softmax operation. 
During training, the gradients calculated by $\mathcal{L}_{\mathtt{CAT}}$ are back-propagated to update the weights in the CAT head (i.e., $\boldsymbol{\gamma}$, $\boldsymbol{\beta}$, $\boldsymbol{W}$) as well as the trainable prompt (i.e., $[\boldsymbol{v}]_1[\boldsymbol{v}]_2...[\boldsymbol{v}]_l$).
Ablation studies in Section \ref{ablation} show that employing two independent cwT layers (one for each modality) is more effective than using a shared cwT layer in the CAT head.

\keypoint{Prompt Tuning with Dual Heads.}
Rather than solely using the designed CAT head to facilitate the preservation of task-shared knowledge during prompt tuning,
our DePT also retains the standard ITM head to learn an alignment of image-text pairs in the original feature space, which is of great importance for establishing better zero-shot generalization on new tasks (as proven in Section \ref{ablation}).
Thus, the overall learning objective of DePT is expressed as:
\begin{equation}
   \mathcal{L} =  \lambda \mathcal{L}_{\mathtt{CAT}}  + (1-\lambda) \mathcal{L}_{\mathtt{ITM}},
   \label{loss_dept}
\end{equation}
where $\lambda$ is a balance weight controlling the relative importance of the two losses.


\keypoint{Test-time Knowledge Fusion for the Base Task.}
At inference, the standard ITM head is used to achieve zero-shot generalization/prediction on new tasks in the original feature space. 
For the base task, our CAT head directly takes the image feature of a testing example as input to predict the in-distribution class label with the linear classifier.
Notably, we can further boost the performance on the base task by simply fusing {base-specific} knowledge in the {CAT} head as well as {task-shard} knowledge in the {ITM} head at inference.
By connecting Eq. (\ref{inf}) and Eq. (\ref{ddd}), the prediction probability of the in-distribution testing example $\boldsymbol{x}$ belonging to the $i$-th class can be computed as:
\begin{equation}
    \mathrm{p}({\boldsymbol{c}_i}| \boldsymbol{x})= \lambda \mathcal{P}_{\mathtt{CAT}}({\boldsymbol{c}_i}| \boldsymbol{x})  + (1-\lambda) \mathcal{P}_{\mathtt{ITM}}({\boldsymbol{c}_i}| \boldsymbol{x}),
    \label{lll}
\end{equation}
where the balance weight $\lambda$ in Eq. (\ref{loss_dept}) is directly used to control the contributions of the two heads for simplification. 
Pytorch-like pseudocode for the implementation of DePT is presented in \underline{\textbf{Sup. Mat.(B)}}.

\begin{table*}[]
\setlength{\abovecaptionskip}{0.1cm}  
\setlength{\belowcaptionskip}{-0.2cm} 
\centering 
\tabcolsep 0.08in
  \footnotesize
\begin{tabular}{l|c|cc|c|ccc}
\hline
\multicolumn{1}{c|}{\multirow{2}{*}{Setting}} & \multicolumn{1}{c|}{\multirow{2}{*}{ITM Head}} & \multicolumn{2}{c|}{CAT Head} & \multicolumn{1}{c|}{{Test-time fusion}} & \multicolumn{3}{c}{Average accuracy over 11 datasets (\%)} \\ 
\cline{3-4} \cline{6-8} \multicolumn{1}{c|}{}  & \multicolumn{1}{c|}{} &  \multicolumn{1}{c}{cwT+LC} & \multicolumn{1}{c|}{cwT+ITM} & \multicolumn{1}{c|}{for the \textit{Base} task} & \multicolumn{1}{c}{Base} & \multicolumn{1}{c}{New} & \multicolumn{1}{c}{H} \\ 
\hline
$\textcircled{1}$ ITM only (\textbf{Baseline})            & \scriptsize{\Checkmark}                                  & \textcolor{lightgray}{\scriptsize{\XSolidBrush}}          &\textcolor{lightgray}{\scriptsize{\XSolidBrush}}  & \textcolor{lightgray}{\scriptsize{\XSolidBrush}}                     &81.50           &69.77            &75.18      \\
\cellcolor{gray!20}$\textcircled{2}$  ITM+CAT (CAT=cwT+LC)   &\cellcolor{gray!20}\scriptsize{\Checkmark}   & \cellcolor{gray!20}\scriptsize{\Checkmark}    &\cellcolor{gray!20}\textcolor{lightgray}{\scriptsize{\XSolidBrush}} & \cellcolor{gray!20}\textcolor{lightgray}{\scriptsize{\XSolidBrush}} &\cellcolor{gray!20}82.14   (\textcolor{blue}{\scriptsize{+0.64}})          &\cellcolor{gray!20}\textbf{71.82}    (\textcolor{blue}{\scriptsize{+\textbf{2.05}}})          &\cellcolor{gray!20}76.63   (\textcolor{blue}{\scriptsize{+1.45}}) \\
\quad \,  \textcolor{gray}{\textit{v1}. Use a {shared} cwT in CAT}  &\scriptsize{\Checkmark}  & \scriptsize{\Checkmark}    &\textcolor{lightgray}{\scriptsize{\XSolidBrush}}           & \textcolor{lightgray}{\scriptsize{\XSolidBrush}}&\textcolor{gray}{82.24  ({\scriptsize{+0.74}})} &\textcolor{gray}{70.85  ({\scriptsize{+1.08}})}         &\textcolor{gray}{76.12  ({\scriptsize{+0.94}})}       \\
\quad \,  \textcolor{gray}{\textit{v2}. Use an ITM classifier in CAT}  &\scriptsize{\Checkmark}  & \textcolor{lightgray}{\scriptsize{\XSolidBrush}}    &\scriptsize{\Checkmark}           & \textcolor{lightgray}{\scriptsize{\XSolidBrush}} &\textcolor{gray}{82.16   ({\scriptsize{+0.66}})}         &\textcolor{gray}{71.31    ({\scriptsize{+1.54}})}          &\textcolor{gray}{76.35   ({\scriptsize{+1.17}})}       \\
\quad \,  \textcolor{gray}{\textit{v3}. Only use image features in CAT}  &\scriptsize{\Checkmark} &\scriptsize{\Checkmark}  & \textcolor{lightgray}{\scriptsize{\XSolidBrush}}              & \textcolor{lightgray}{\scriptsize{\XSolidBrush}} & \textcolor{gray}{81.11 ({\scriptsize{- 0.39}})}   & \textcolor{gray}{70.93 ({\scriptsize{+1.16}})}   & \textcolor{gray}{75.68  ({\scriptsize{+0.50}})}   \\
\cellcolor{gray!20}$\textcircled{3}$ ITM+CAT+Fusion (\textbf{Our DePT})   & \cellcolor{gray!20}\scriptsize{\Checkmark}   &\cellcolor{gray!20}\scriptsize{\Checkmark}   & \cellcolor{gray!20}\textcolor{lightgray}{\scriptsize{\XSolidBrush}}& \cellcolor{gray!20}\scriptsize{\Checkmark}   & \cellcolor{gray!20}\textbf{83.66}  (\textcolor{blue}{\scriptsize{+\textbf{2.16}}})      &  \cellcolor{gray!20}\textbf{71.82}  (\textcolor{blue}{\scriptsize{+\textbf{2.05}}})      & \cellcolor{gray!20}\textbf{77.29}  (\textcolor{blue}{\scriptsize{+\textbf{2.11}}})     \\   
\hline
\end{tabular}
\caption{Ablation study for the designed components of DePT. The baseline method is CoOp \cite{zhou2022learning}, and the average accuracy on 11 datasets are reported. The metric “H” indicates the harmonic mean \cite{zhou2022conditional} of {base-task} and {new-task} accuracies. “LC”: Linear Classifier.}
\label{table-ab}
\end{table*}

\section{Experiments}
\label{sec_exp}
In this section, we first present ablation studies to analyze the impacts of different factors on DePT.
Next, we validate the flexibility and effectiveness of DePT by applying it to several baseline schemes. 
We start with an introduction of the experimental setup below.

\subsection{Experimental Setup}

\keypoint{Baselines.}
We apply our DePT to a broad spectrum of baseline approaches, including the \textit{visual} prompt tuning method VPT \cite{jia2022vpt}, \textit{textual} prompt tuning methods CoOp \cite{zhou2022learning}, CoCoOp \cite{zhou2022conditional} and KgCoOp \cite{yao2023visual}, and \textit{multi-model} prompt tuning methods MaPLe \cite{khattak2023maple}, PromptSRC \cite{Khattak_2023_ICCV}. 


\keypoint{Datasets}. 
We conduct experiments on several datasets from diverse sources. Concretely, for the settings of \textit{{base-to-new generalization}} and \textit{cross-dataset generalization}, we use \textbf{11} datasets: ImgNet \cite{deng2009imagenet}, Caltech \cite{fei2004learning}, OxfordPets \cite{parkhi2012cats}, StanfordCars \cite{krause20133d}, Flowers \cite{nilsback2008automated},
Food101 \cite{bossard2014food}, FGVCAircraft \cite{maji2013fine}, EuroSAT \cite{helber2019eurosat}, UCF101 \cite{soomro2012ucf101}, DTD \cite{cimpoi2014describing}, and SUN397 \cite{xiao2010sun};
for the \textit{domain generalization} setting, we use ImgNet as the source domain (i.e. the base task), and its four variants ImgNet-V2 \cite{recht2019imagenet}, ImgNet-Sketch \cite{wang2019learning}, ImgNet-A \cite{gao2022generating} and ImgNet-R \cite{hendrycks2021many} as target domains (i.e. new tasks).

\keypoint{Implementation Details.}
Our implementations are based on the open-source repository of MaPLe \cite{khattak2023maple}\footnote{{https://github.com/muzairkhattak/multimodal-prompt-learning}}. 
For each baseline method, we use the same experimental setup (e.g., feature backbone, prompt length and learning rate) as used in the original implementation. 
For DePT, the learning rate for updating the parameters in the devised CAT head is set to $6.5 \times \delta$, where $\delta$ is the adopted learning rate of each baseline for prompt tuning.
We adjust the value of $\lambda$ and the training epoch for our DePT in ablation studies.
The above hyperparameters are fixed across all datasets.
Unless stated otherwise, the base task is constructed as a many-way 16-shot task.
We report {base-task} and {new-task} accuracies as well as their harmonic-mean (H) \cite{zhou2022conditional} averaged over 3 runs to compare the performance of different methods.
Code: \url{https://github.com/Koorye/DePT}.

\subsection{Ablation Studies}
\begin{figure}[t]
\setlength{\abovecaptionskip}{0.05cm}  
\setlength{\belowcaptionskip}{-0.25cm} 
\centering
\includegraphics[width=1.0\linewidth]{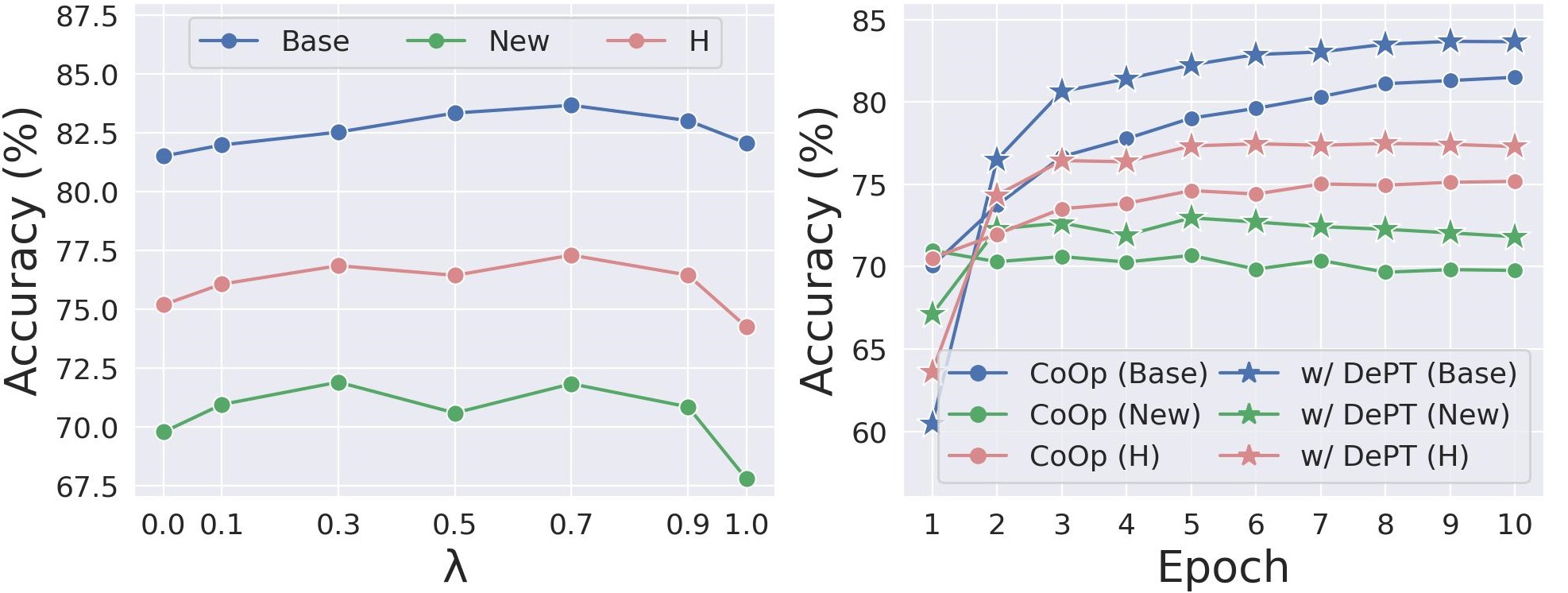} 
\caption{\textbf{Left}: Impact of the balance weight $\lambda$ in Eq. (\ref{loss_dept})/(\ref{lll}) on DePT. 
\textbf{Right}: Performance of DePT at different training epochs.} 
\label{hyper}
\end{figure}

\begin{figure}[t]
\setlength{\abovecaptionskip}{0.1cm}  
\setlength{\belowcaptionskip}{-0.55cm} 
\centering
\includegraphics[width=1.0\linewidth]{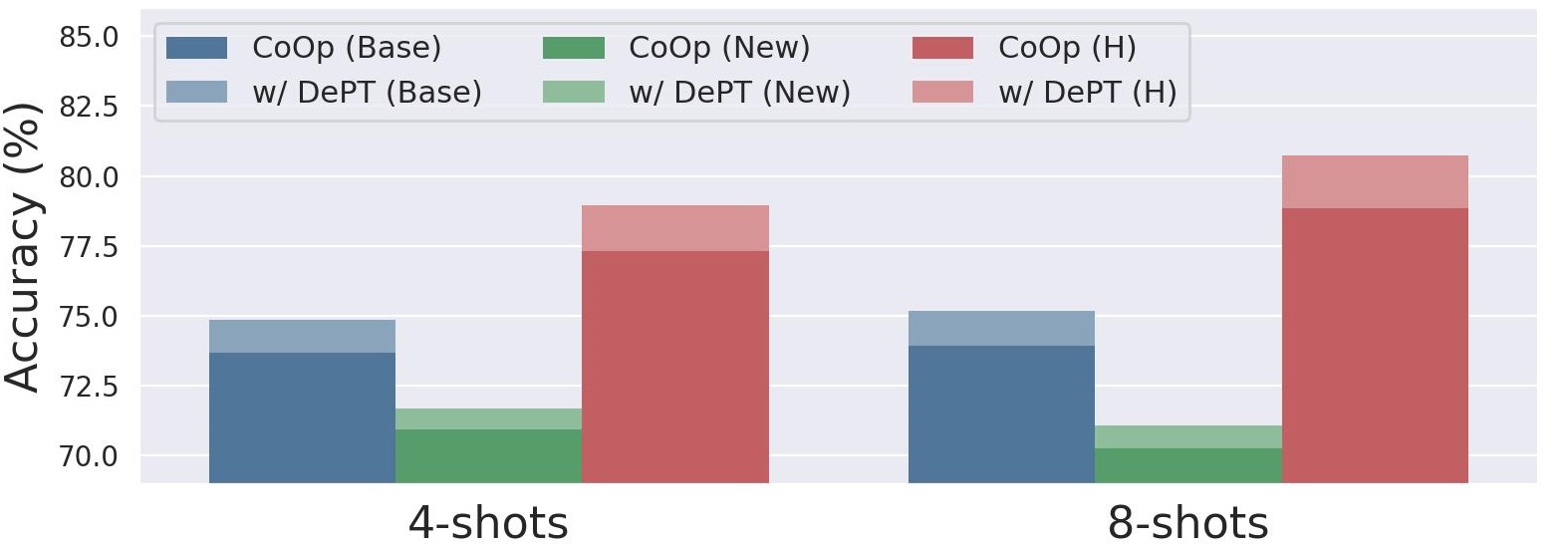} 
\caption{Robustness of DePT under different shots.}
\label{bar}
\end{figure}

Here, we first conduct an ablative analysis of the designed components of DePT in Table \ref{table-ab}. 
Then, we investigate the impact of the balance weight $\lambda$ on DePT in Figure \ref{hyper} (\textbf{Left}). 
Next, we scrutinize the performance of DePT on different training epochs in Figure \ref{hyper} (\textbf{Right}).
Finally, we validate the robustness of DePT under different shots in Figure \ref{bar}.
We perform experiments using the baseline method CoOp \cite{zhou2022learning} in the base-to-new generalization setting, results averaged on the \textbf{11} datasets are reported.

\label{ablation}
\keypoint{Effectiveness of the Devised Components in DePT.}
Our DePT contains two key components, including a plug-and-play CAT head for capturing base-specific knowledge in an isolated feature space, as well as a test-time knowledge fusion strategy for exploring both base-specific and task-shard knowledge to improve the performance on the base task. 
We conduct component-wise analysis on the two components by progressively adding one of them to the baseline method CoOp \cite{zhou2022learning} in Table \ref{table-ab}, where the results are averaged over 11 datasets.
From $\textcircled{1}$ and $\textcircled{2}$ in the table, we observe that integrating our CAT head with the standard ITM head for prompt tuning improves both base-task and new-task accuracies of the baseline method, achieving a clear enhancement of the harmonic-mean (by \textbf{1.45}\%).
Notably, $\textcircled{2}$ outperforms $\textcircled{1}$ by up to \textbf{2.05}\% in terms of new-task accuracy, which demonstrates the effectiveness of our CAT head in facilitating the preservation of task-shared knowledge during prompt tuning.
Besides, we also compare the CAT head with its three variants.
Concretely, we replace the two independent cwT layers (one for each modality) with a shared cwT layer in \textit{v1}, we replace the linear classifier with an ITM classifier in \textit{v2}, and we only feed image features to the liner classifier in \textit{v3} (more details are in \underline{\textbf{Sup. Mat.(C)}}).
As shown, all the three variants underperform our designed CAT head.
What is noteworthy is that directly appending a standard ITM classifier in the cwT-transformed feature space also considerably improves the performance of the baseline on new tasks (see \textit{v2}), showing the effectiveness of the CAT head for decoupling base-specific knowledge and task-shared knowledge during prompt tuning.
Besides, we see that using only image features in the CAT head damages the performance on the base task (see \textit{v3}).
This is possibly due to that relying on a limited number of examples for model optimization, the parameters in the CAT head may overfit to the training data of the base task when the gradients of $\mathcal{L}_{\mathtt{CAT}}$ can not be back-propagated to the text encoder to optimize the parameters of the prompt.
What's more, by simply fusing base-specific knowledge and task-shard knowledge in the two heads at inference, the performance on the base task can be enhanced considerably, achieving an absolute gain of \textbf{2.16}\% in accuracy, as shown in $\textcircled{3}$.

\keypoint{Impact of the Balance Weight $\lambda$ on DePT.}
In the proposed DePT, we employ the balance weight $\lambda$ to control the relative importance of the standard ITM head and our devised CAT head in Eq. (\ref{loss_dept})/(\ref{lll}).
It is necessary to investigate the impact of $\lambda$ on the performance of DePT.
To this end, we set $\lambda$ to the values of $\{0.0, 0.1, 0.2, ..., 1.0\}$, and report the average testing results on the 11 datasets in Figure \ref{hyper} (\textbf{Left}).
Overall, the performance of DePT gradually increases as the $\lambda$ value grows from 0.0 to 0.7, after which the performance of DePT gradually decreases and reaches the lowest value when $\lambda$=1.0.
In particular, when $\lambda$=0.7 DePT establishes the best performance on both base and new tasks.
What is noteworthy is that when $\lambda$=1.0, i.e., only our CAT head is used for training, the performance of DePT on new tasks sharply decreases, which suggests that retaining the ITM head to learn an alignment of positive image-text features in the original feature space is of great importance for achieving better zero-shot prediction performance on new tasks.

\keypoint{Performance of DePT at Different Training Epochs.}
In Figure \ref{hyper} (\textbf{Right}), we report the obtained results of the baseline method w/ or w/o our DePT at different training epochs.
As can be observed, our DePT consistently improves the baseline method after epoch 2, in terms of base-task, new-task, and harmonic mean (H) accuracies. 
One possible reason for the failure case at epoch 1 is that the weights in the CAL head (i.e., $\boldsymbol{\gamma}$, $\boldsymbol{\beta}$, $\boldsymbol{W}$) are initialized randomly, thus it is difficult for the CAL head to fully capture base-specific knowledge with only one training epoch.
We also see the baseline fails to address the BNT problem during prompt tuning -- overall, the accuracy of the baseline on new tasks decreases as its performance on base tasks increases from epoch 1 to epoch 10. 
It is obvious that DePT mitigates the BNT problem effectively.
The performance of the baseline method and DePT is saturated at epoch 10.

\keypoint{Robustness of DePT under Different Shots.}
All the aforementioned results are obtained on many-way 16-shot base tasks -- for every base task, 16 training examples are sampled from each class for prompt tuning.
It is interesting to further scrutinize the robustness of our DePT under different shots.
To achieve this, we set the shots to $\{4,8,16\}$, and report the average testing results of the baseline method w/ or w/o our DePT on the 11 datasets in Figure \ref{bar}.
As can be observed, our DePT consistently improves the baseline method across all 4-shot, 8-shot, and 16-shot settings, in terms of base-task, new-task, and harmonic mean (H) accuracies, 
showing the robustness of DePT under few shots. 
In the following section, we follow \cite{zhou2022learning,yao2023visual,zhou2022conditional,khattak2023maple} to evaluate methods in the 16-shot setting.

\keypoint{Computational Cost.} The computational cost of our DePT is shown in Table \ref{cost}. 
As observed, the additional computational cost is low (even \textit{negligible}) compared to the performance improvement established by DePT.
\begin{table}[t]
\setlength{\abovecaptionskip}{0.1cm}  
\setlength{\belowcaptionskip}{-0.4cm} 
		\centering
		 \tabcolsep 0.052in
		\scriptsize
		\begin{tabular}{l|cccc|c}
		\hline
		  Method & Learnable para. & Train.time & Infer.time & Memory & H (\textit{avg}) \\
            \hline
            CoOp& 8K &105min &2.78ms &11264MB  &75.18\\
               +\textbf{DePT}&+(2+N/2)K  &\color{blue}{+0min}\color{black} &\color{blue}{+0ms}\color{black} &\color{blue}+2MB\color{black} &\textbf{77.29} (\color{blue}{+2.11}\color{black})\\
            \hline            
		\end{tabular}
		\caption{{Computational cost of DePT}. “N”: the num of classes in the base task, “ms”: millisecond per image. Experiments are performed on a V100 GPU.}
		\label{cost}
\end{table}

 \begin{table*}[htbp]
\setlength{\abovecaptionskip}{0.1cm}  
\setlength{\belowcaptionskip}{-0.1cm} 
  \centering 
  \tabcolsep 0.12in
  \footnotesize
    \begin{tabular}{l|ccc|ccc|ccc|ccc}
      \hline
     \multicolumn{1}{l|}{\multirow{2}{*}{\makecell[c]{{Method}}}}  & \multicolumn{3}{c|}{{{\textbf{Avg over 11 datasets}}}}  & \multicolumn{3}{c|}{\cellcolor{blue!0}{ImageNet}} & \multicolumn{3}{c|}{\cellcolor{gray!0}{Caltech101}}  & \multicolumn{3}{c}{\cellcolor{gray!0}{OxfordPets}}  \\
    \cline{2-13} 
    &  \multicolumn{1}{c}{Base} & \multicolumn{1}{c}{New} & \multicolumn{1}{c|}{H} &  \multicolumn{1}{c}{Base} & \multicolumn{1}{c}{New} & \multicolumn{1}{c|}{H}  & \multicolumn{1}{c}{Base} & \multicolumn{1}{c}{New} & \multicolumn{1}{c|}{H}  & \multicolumn{1}{c}{Base} & \multicolumn{1}{c}{New} & \multicolumn{1}{c}{H} \\
     \hline
     {CoOp} \cite{zhou2022learning}  &81.50 &69.77 &75.18 &76.57 &69.97 &73.12 &98.17 &\textbf{94.83} &\textbf{96.47}  &\textbf{95.57} &97.53 &\textbf{96.54}\\
     \cellcolor{gray!20}\textbf{+DePT}  &\cellcolor{gray!20}{\textbf{\textcolor{blue}{83.66}}} &\cellcolor{gray!20}{\textbf{\textcolor{blue}{71.82}}} &\cellcolor{gray!20}{\textbf{\textcolor{blue}{77.29}}} &\textbf{77.13} &\textbf{70.10} &\textbf{73.45} &\textbf{98.33} &94.33 &96.29 &94.70 &\textbf{97.63} &96.14\\
    \hline
         {CoCoOp} \cite{zhou2022conditional} &81.18 &72.18 &76.42 &75.90 &\textbf{70.73} &73.23 &97.70 &93.20 &95.40  &\textbf{94.93} &\textbf{97.90} &\textbf{96.39}\\
     \cellcolor{gray!20}\textbf{+DePT}   &\cellcolor{gray!20}{\textbf{\textcolor{blue}{83.80}}} &\cellcolor{gray!20}{\textbf{\textcolor{blue}{72.89}}} &\cellcolor{gray!20}{\textbf{\textcolor{blue}{77.97}}} &\textbf{76.87} &70.47 &\textbf{73.53} &\textbf{98.37} &\textbf{93.87} &\textbf{96.06} &94.03 &97.20 &95.59\\
    \hline
         {KgCoOp} \cite{yao2023visual} &80.93 &73.88 &77.25 &76.17 &\textbf{70.53} &73.24 &97.87 &94.03 &95.91 &\textbf{95.47} &\textbf{97.80} &\textbf{96.62}\\
     \cellcolor{gray!20}\textbf{+DePT}   &\cellcolor{gray!20}{\textbf{\textcolor{blue}{83.62}}} &\cellcolor{gray!20}{\textbf{\textcolor{blue}{75.04}}} &\cellcolor{gray!20}{\textbf{\textcolor{blue}{79.10}}} &\textbf{77.03} &70.13 &\textbf{73.42} &\textbf{98.30} &\textbf{94.60} &\textbf{96.41} &94.33 &97.23 &95.76\\
    \hline
         {MaPLe} \cite{khattak2023maple} &83.54 &73.76 &78.35 &77.23 &69.63 &73.24 &98.30 &93.70 &95.94 &\textbf{95.17} &97.77 &\textbf{96.45}\\
     \cellcolor{gray!20}\textbf{+DePT} &\cellcolor{gray!20}{\textbf{\textcolor{blue}{84.85}}} &\cellcolor{gray!20}{\textbf{\textcolor{blue}{74.82}}} &\cellcolor{gray!20}{\textbf{\textcolor{blue}{79.52}}}&\textbf{77.87} &\textbf{70.23} &\textbf{73.85} &\textbf{98.53} &\textbf{95.03} &\textbf{96.75} &95.03 &\textbf{97.83} &96.41\\
    \hline
         {VPT} \cite{jia2022vpt} &82.11 &71.73 &76.57 &75.90 &68.10 &71.79 &98.03 &94.30 &96.13 &95.13 &\textbf{96.47} &95.80 \\
        \cellcolor{gray!20}\textbf{+DePT} &\cellcolor{gray!20}\textbf{\textcolor{blue}{85.28}} &\cellcolor{gray!20}\textbf{\textcolor{blue}{73.96}} &\cellcolor{gray!20}\textbf{\textcolor{blue}{79.22}} &\textbf{78.40} &\textbf{68.90} &\textbf{73.34} &\textbf{98.67} &\textbf{94.33} &\textbf{96.45} &\textbf{95.50} &96.33 &\textbf{95.91} \\
    \hline
         {PromptSRC} \cite{Khattak_2023_ICCV} &84.21 &75.75 &79.76 &77.63 &70.23 &73.75 &98.10 &93.87 &95.94 &95.27 &97.23 &96.24 \\
        \cellcolor{gray!20}\textbf{+DePT} &\cellcolor{gray!20}\textbf{\textcolor{blue}{85.19}} &\cellcolor{gray!20}\textbf{\textcolor{blue}{76.17}} &\cellcolor{gray!20}\textbf{\textcolor{blue}{80.43}} &\textbf{78.20} &\textbf{70.27} &\textbf{74.02} &\textbf{98.57} &\textbf{94.10} &\textbf{96.28} &\textbf{95.43} &\textbf{97.33} &\textbf{96.37} \\
     \hline
     \multicolumn{1}{l|}{\multirow{2}{*}{\makecell[c]{{Method}}}}  & \multicolumn{3}{c|}{\cellcolor{gray!0}{StanfodCars}}  & \multicolumn{3}{c|}{\cellcolor{gray!0}{Flowers102}} & \multicolumn{3}{c|}{\cellcolor{gray!0}{Food101}} & \multicolumn{3}{c}{\cellcolor{gray!0}{FGVCAircraft}}    \\
    \cline{2-13} 
    &  \multicolumn{1}{c}{Base} & \multicolumn{1}{c}{New} & \multicolumn{1}{c|}{H} &  \multicolumn{1}{c}{Base} & \multicolumn{1}{c}{New} & \multicolumn{1}{c|}{H}  &  \multicolumn{1}{c}{Base} & \multicolumn{1}{c}{New} & \multicolumn{1}{c|}{H}  &  \multicolumn{1}{c}{Base} & \multicolumn{1}{c}{New} & \multicolumn{1}{c}{H}  \\
     \hline
     {CoOp} \cite{zhou2022learning} &74.30 &72.10 &73.18 &97.07 &\textbf{74.33} &\textbf{84.19} &{90.43} &90.97 &90.70 &31.70 &17.30 &22.38\\
     \textbf{+DePT} &\textbf{79.67} &\textbf{72.40} &\textbf{75.86} &\textbf{98.20} &72.00 &83.08 &\textbf{90.43} &\textbf{91.33} &\textbf{90.88} &\textbf{42.53} &\textbf{22.53} &\textbf{29.46}\\
    \hline
         {CoCoOp} \cite{zhou2022conditional} &70.77 &72.50 &71.62 &95.03 &69.07 &80.00 &\textbf{90.57} &91.20 &\textbf{90.88} &35.63 &\textbf{32.70} &34.10\\
     \textbf{+DePT}  &\textbf{79.87} &\textbf{73.33} &\textbf{76.46} &\textbf{98.33} &\textbf{72.57} &\textbf{83.51} &90.30 &\textbf{91.30} &90.80 &\textbf{43.07} &31.30 &\textbf{36.25}\\
    \hline
         {KgCoOp} \cite{yao2023visual} &71.13 &74.67 &72.86 &95.90 &74.83 &84.07 &90.47 &\textbf{91.60} &91.03 &35.10 &\textbf{35.20} &35.15\\
     \textbf{+DePT}  &\textbf{79.13} &\textbf{75.47} &\textbf{77.26} &\textbf{98.00} &\textbf{76.37} &\textbf{85.84} &\textbf{90.50} &\textbf{91.60} &\textbf{91.05} &\textbf{43.20} &34.83 &\textbf{38.57}\\
    \hline
         {MaPLe} \cite{khattak2023maple} &76.30 &\textbf{72.53} &74.37 &97.23 &72.07 &82.78 &90.30 &\textbf{91.53} &90.91 &40.57 &\textbf{36.47} &\textbf{38.31}\\
     \textbf{+DePT}   &\textbf{80.93} &71.73 &\textbf{76.06} &\textbf{98.03} &\textbf{73.17} &\textbf{83.79} &\textbf{90.33} &\textbf{91.53} &\textbf{90.93} &\textbf{44.53} &32.80 &37.78\\
    \hline
         {VPT} \cite{jia2022vpt} &71.63 &\textbf{72.20} &71.92 &95.93 &70.37 &81.18 &89.80 &90.37 &90.08 &35.90 &30.37 &32.90 \\
        \textbf{+DePT} &\textbf{82.13} &72.17 &\textbf{76.83} &\textbf{98.17} &\textbf{73.20} &\textbf{83.86} &\textbf{90.27} &\textbf{91.03} &\textbf{90.65} &\textbf{45.30} &\textbf{31.87} &\textbf{37.41} \\
    \hline
         {PromptSRC} \cite{Khattak_2023_ICCV} &78.37 &74.97 &76.63 &97.90 &76.97 &86.18 &90.63 &91.53 &91.08 &42.53 &\textbf{36.87} &39.50 \\
        \textbf{+DePT} &\textbf{80.80} &\textbf{75.00} &\textbf{77.79} &\textbf{98.40} &\textbf{77.10} &\textbf{86.46} &\textbf{90.87} &\textbf{91.57} &\textbf{91.22} &\textbf{45.70} &36.73 &\textbf{40.73} \\
\hline
     \multicolumn{1}{l|}{\multirow{2}{*}{\makecell[c]{{Method}}}}  & \multicolumn{3}{c|}{\cellcolor{gray!0}{SUN397}}  & \multicolumn{3}{c|}{\cellcolor{gray!0}{DTD}} & \multicolumn{3}{c|}{\cellcolor{gray!0}{EuroSAT}}  & \multicolumn{3}{c}{\cellcolor{gray!0}{UCF101}}  \\
    \cline{2-13} 
    &  \multicolumn{1}{c}{Base} & \multicolumn{1}{c}{New} & \multicolumn{1}{c|}{H} &  \multicolumn{1}{c}{Base} & \multicolumn{1}{c}{New} & \multicolumn{1}{c|}{H}  &  \multicolumn{1}{c}{Base} & \multicolumn{1}{c}{New} & \multicolumn{1}{c|}{H}  &  \multicolumn{1}{c}{Base} & \multicolumn{1}{c}{New} & \multicolumn{1}{c}{H}   \\
     \hline
     {CoOp} \cite{zhou2022learning} &81.13 &\textbf{76.07} &78.52  &79.33 &49.70 &61.11 &\textbf{89.35} &57.30 &69.82 &83.87 &69.80 &76.19\\
     \textbf{+DePT}  &\textbf{82.37} &75.07 &\textbf{78.55} &\textbf{83.20} &\textbf{56.13} &\textbf{67.04} &88.27 &\textbf{66.27} &\textbf{75.70} &\textbf{85.43} &\textbf{72.17} &\textbf{78.24}\\
    \hline
         {CoCoOp} \cite{zhou2022conditional} &79.50 &76.27 &77.85 &77.37 &52.97 &62.88 &87.97 &63.63 &73.85 &82.33 &72.40 &77.05\\
     \textbf{+DePT}  &\textbf{82.20} &\textbf{76.73} &\textbf{79.37} &\textbf{82.77} &\textbf{55.40} &\textbf{66.37} &\textbf{90.27} &\textbf{66.87} &\textbf{76.82} &\textbf{85.70} &\textbf{72.80} &\textbf{78.73}\\
    \hline
         {KgCoOp} \cite{yao2023visual} &80.40 &77.30 &78.82 &78.27 &57.93 &66.58 &85.77 &63.40 &72.91 &83.73 &75.40 &79.35\\
     \textbf{+DePT}  &\textbf{82.33} &\textbf{77.80} &\textbf{80.00} &\textbf{82.20} &\textbf{59.13} &\textbf{68.78} &\textbf{89.03} &\textbf{71.07} &\textbf{79.04} &\textbf{85.80} &\textbf{77.23} &\textbf{81.29}\\
    \hline
         {MaPLe} \cite{khattak2023maple} &81.93 &\textbf{76.50} &79.12  &81.93 &58.20 &68.06 &\textbf{94.67} &66.73 &78.28 &85.30 &76.23 &80.51\\
     \textbf{+DePT}  &\textbf{82.90} &76.40 &\textbf{79.52}&\textbf{83.87} &\textbf{59.93} &\textbf{69.91} &94.43 &\textbf{76.23} &\textbf{84.36} &\textbf{86.87} &\textbf{78.10} &\textbf{82.25}\\
    \hline
         {VPT} \cite{jia2022vpt} &79.50 &76.17 &77.80 &80.90 &52.73 &63.85 &\textbf{95.83} &65.03 &77.48 &84.63 &72.90 &78.33 \\
        \textbf{+DePT} &\textbf{83.03} &\textbf{77.77} &\textbf{80.31} &\textbf{85.07} &\textbf{56.60} &\textbf{67.97} &93.77 &\textbf{76.30} &\textbf{84.14} &\textbf{87.73} &\textbf{75.10} &\textbf{80.93} \\
    \hline
         {PromptSRC} \cite{Khattak_2023_ICCV} &82.63 &\textbf{78.97} &80.76 &83.43 &\textbf{62.53} &\textbf{71.49} &92.80 &72.07 &81.13 &87.03 &\textbf{78.07} &82.31 \\
        \textbf{+DePT} &\textbf{83.27} &\textbf{78.97} &\textbf{81.06} &\textbf{84.80} &61.20 &71.09 &\textbf{93.23} &\textbf{77.90} &\textbf{84.88} &\textbf{87.73} &77.70 &\textbf{82.46} \\
     \hline
    \end{tabular}
    \caption{{Base-to-new generalization} performance of six baselines w/ or w/o our DePT on 11 datasets. }
  \label{table—B2N}
\end{table*}

\begin{table*}[htbp]
\setlength{\abovecaptionskip}{0.1cm}  
\setlength{\belowcaptionskip}{-0.4cm} 
  \centering 
  \tabcolsep 0.038in
  \footnotesize
    \begin{tabular}{l|c|ccccccccccc}
      \hline
     \multicolumn{1}{l|}{\multirow{2}{*}{\makecell[c]{{Method}}}}  & {{Source}}  & \multicolumn{10}{c}{{Target}}   \\
    \cline{3-13} 
    & \multicolumn{1}{c|}{(ImgNet)} & {\, {\textbf{Avg}} \, } & Caltech101 & {OxfordPets} & {StanfCars} & {Flowers102} & {Food101}& {FGVCAircraft}& {SUN397}& {DTD}& {EuroSAT}& {UCF101} \\
     \hline
     {CoOp} \cite{zhou2022learning} &71.80 &64.40  &\textbf{93.97} &89.60 &64.60 &69.13 &85.47 &20.70 &65.70 &43.07 &44.50 &67.23  \\
     \cellcolor{gray!20}\textbf{+DePT} &\cellcolor{gray!20}\textbf{\textcolor{blue}{72.63}} &\cellcolor{gray!20}\textbf{\textcolor{blue}{65.02}}  &93.30 &\textbf{90.00} &\textbf{65.53} &\textbf{70.50} &\textbf{85.97} &\textbf{21.90} &\textbf{66.07} &\textbf{43.17} &\textbf{44.97} &\textbf{68.80} \\
    \hline
         {CoCoOp} \cite{zhou2022conditional} &71.17 &65.73  &\textbf{94.30} &\textbf{90.80} &65.53 &71.80 &86.13 &22.83 &\textbf{67.73} &\textbf{45.57} &43.47 &69.10  \\
     \cellcolor{gray!20}\textbf{+DePT} &\cellcolor{gray!20}\textbf{\textcolor{blue}{72.77}}  &\cellcolor{gray!20}\textbf{\textcolor{blue}{65.88}} &94.10 &90.63 &\textbf{66.23} &\textbf{72.17} &\textbf{86.27} &\textbf{22.90} &67.30 &45.50 &\textbf{44.17} &\textbf{69.53} \\
    \hline
         {KgCoOp} \cite{yao2023visual} &71.17 &65.06   &94.17 &89.70 &64.77 &70.30 &\textbf{86.47} &22.43 &\textbf{66.83} &44.43 &\textbf{43.53} &68.00 \\
     \cellcolor{gray!20}\textbf{+DePT}   &\cellcolor{gray!20}\textbf{\textcolor{blue}{72.77}} &\cellcolor{gray!20}\textbf{\textcolor{blue}{65.55}} &\textbf{94.23} &\textbf{90.03} &\textbf{65.57} &\textbf{70.57} &86.37 &\textbf{23.27} &66.67 &\textbf{45.97} &\textbf{43.53} &\textbf{69.30}\\
    \hline
         {MaPLe} \cite{khattak2023maple}  &72.47 &64.17  &\textbf{92.97} &\textbf{90.20} &63.97 &70.03 &84.83 &23.23 &66.00 &43.23 &40.03 &67.23 \\
    \cellcolor{gray!20} \textbf{+DePT} &\cellcolor{gray!20}\textbf{\textcolor{blue}{73.27}}  &\cellcolor{gray!20}\textbf{\textcolor{blue}{64.56}} &92.53 &90.10 &\textbf{64.60} &\textbf{70.10} &\textbf{85.57} &\textbf{23.63} &\textbf{66.40} &\textbf{45.03} &\textbf{40.13} &\textbf{67.53} \\
    \hline 
        {VPT} \cite{jia2022vpt} &70.80 &62.61   &\textbf{91.67} &\textbf{90.03} &62.47 &66.03 &81.70 &\textbf{24.07} &65.27 &44.27 &35.77 &64.83 \\
    \cellcolor{gray!20} \textbf{+DePT}  &\cellcolor{gray!20}\textbf{\textcolor{blue}{71.97}} &\cellcolor{gray!20}\textbf{\textcolor{blue}{63.01}} &91.30 &\textbf{90.03} &\textbf{62.63} &\textbf{66.77} &\textbf{83.03} &23.73 &\textbf{65.57} &\textbf{44.57} &\textbf{37.03} &\textbf{65.40}  \\
    \hline 
        {PromptSRC} \cite{Khattak_2023_ICCV} &71.33 &65.71 &93.77 &\textbf{90.40} &65.77 &70.80 &\textbf{86.30} &23.67 &66.93 &46.07 &44.23 &\textbf{69.20} \\
    \cellcolor{gray!20} \textbf{+DePT} &\cellcolor{gray!20}\textcolor{blue}{\textbf{71.60}} &\cellcolor{gray!20}\textcolor{blue}{\textbf{66.02}} &\textbf{93.80} &90.13 &\textbf{66.00} &\textbf{70.93} &86.27 &\textbf{24.30} &\textbf{67.23} &\textbf{46.60} &\textbf{45.83} &69.10  \\
     \hline 
    \end{tabular}
    \caption{Cross-dataset generalization performance of six baselines w/ or w/o our DePT  on 11 datasets.}
  \label{table—CD}
\end{table*}

\subsection{Experimental Results}
\label{expres}
In this part, we demonstrate the flexibility and effectiveness of DePT in the {base-to-new generalization} (in Table \ref{table—B2N}) and {cross-dataset generalization} (in Table \ref{table—CD}) settings. 

\keypoint{Base-to-New Generalization.}
The {base-to-new generalization} setting evaluates whether the models learned on base tasks can generalize to new tasks with unseen classes, i.e., there is a \textit{category shift} between base and new tasks. 
Following the baseline methods, for each dataset, we first construct a base task and a new task by equally dividing the dataset into two sets of classes, then we perform prompt tuning on the base task and test the learned model on both the base and new tasks.
Table \ref{table—B2N} presents the base-to-new generalization performance of the six baselines w/ or w/o our DePT framework over 11 datasets. 
From the average results in the table, we observe a tradeoff between base-task and new-task accuracies for most of the baseline methods, e.g., CoCoOp outperforms CoOp on new tasks but underperforms CoOp on base tasks.
Notably, DePT consistently improves the performance of all baselines without performance tradeoffs on base and new tasks. Specifically, DePT improves each baseline in terms of all base-task, new-task and harmonic-mean accuracies.
From the results on 11 datasets, we also observe some failure cases, e.g., on the OxfordPets dataset, DePT fails to bring clear performance gains on most baseline methods.
Possible reasons are as following.
\textbf{1)} The optimal hyperparameters of DePT for different datasets and baselines are quite different, while we fix them across all datasets and baselines. 
\textbf{2)} When the \textit{category shift} between downstream tasks and the upstream data for model (i.e. CLIP) pretraining is minimal, the advantages of our DePT for task adaptation become less significant.

\keypoint{Cross-Dataset Generalization.}
The cross-dataset generalization setting evaluates whether the model learned on the source dataset can generalize to unseen target datasets, i.e., there is a \textit{distribution shift} between base and new tasks. 
In this experiment, we follow the baselines to regard ImgNet as the source dataset and the other 10 datasets as target datasets. Table \ref{table—CD} presents the performance of the six baselines w/ or w/o DePT on the 11 datasets. 
As can be seen, our DePT consistently improves the accuracy on the source dataset for all baselines, without compromising the performance on 10 target datasets in most cases. 
Notably, 
on average our DePT consistently enhances the performance of all baselines on both the source and target datasets, suggesting DePT is robust to the distribution shift.
Moreover, we see that the previous state-of-the-art method MaPLe establishes the best base-to-new generalization performance among the six baseline methods in Table \ref{table—B2N}, but in Table \ref{table—CD} it achieves inferior cross-dataset generalization performance to CoCoOp and KgCoOp. 
This is probably due to that without decoupling base-specific knowledge and task-shared knowledge during prompt tuning, the learned prompts in both the image encoder and the text encoder for MaPLe are not generalizable enough under distribution shift.

\keypoint{Domain Generalization.} We also evaluate DePT in the domain generalization setting using the six baseline methods in \underline{\textbf{Sup.Mat (D)}}, where DePT still maintains the advantages as in previous settings. 

\keypoint{Effectiveness of DePT on Adapter Tuning Methods.} We also apply DePT to the adapter tuning method CLIP-adapter \cite{gao2021clipad} in \underline{\textbf{Sup.Mat (E)}}, where DePT still maintains the advantages as on the six prompt tuning methods.

\section{Related Work}
\keypoint{Vision-Language Pre-training.} 
Deep learning algorithms have been reported to exhibit or even surpass human-level performance on computer vision and natural language processing tasks \cite{shao2024collaborative,shao2023attention,shao2023fads,fu2021meta,fu2023styleadv,zhang2023global,zhang2022progressive,jiangxun_CVPR2022,jiangxun_TNNLS2022}.
By modeling a connection of image-text pairs, large vision-language pre-trained models (VLPMs) have shown strong zero-shot generalization on various downstream tasks.
Generally, VLPMs leverage three types of pre-text tasks for modeling the semantic correspondence between the vision and language modalities, including \textbf{1)} image-text matching \cite{kim2021vilt,jia2021scaling}, \textbf{2)} contrastive learning \cite{huo2021wenlan,li2021align,DMG_ICML,mo2024selfsupervised}, and \textbf{3)} masked vision/language prediction \cite{kim2021vilt,lu2019vilbert}. 
Beyond image domain, 3D-VLP \cite{jin2023context} initially achieves VLPM on sparse and irregular point clouds using a novel mutual mask modeling.
In this work, we mainly focus on VLPMs establishing image-text alignment with contrastive learning, motivated by their excellent generalization ability to downstream tasks.
For example, after seeing 400 million text-image pairs, CLIP \cite{radford2021clip} learns an alignment between visual and textual features output by an image encoder and a text encoder respectively. 
Beyond recognition \cite{zhou2022conditional,khattak2023maple,fang2023pros,zhang2022tip}, CLIP also demonstrates great potential for other downstream applications, such as image manipulation \cite{wang2022clip,patashnik2021styleclip}, video-text retrieval \cite{ma2022x,zhu2023complementarity,cheng2021improving}, and dense prediction  \cite{rao2022denseclip,zhou2022extract}.

\keypoint{Task Adaptation on VLPMs.} 
The remarkable success of VLPMs have brought new light but also pose a new question: how to efficiently adapt the knowledge from VLPMs to different downstream tasks? 
The most direct solution is \textit{full-finetuning}, which fixes the architecture of VLPMs and tunes all the parameters on the target task. While the results are impressive, this line of work becomes prohibitively expensive with the ever-increasing size of parameters of VLPMs. 
To remedy this, \textit{partial-finetuning} has been proposed to update only a small number of extra parameters (a.k.a. \textit{adapters}) while keeping most pre-trained parameters frozen. Representative schemes are Adapters \cite{houlsby2019adapters}, CLIP-adapter \cite{gao2021clipad}, LoRA \cite{hu2021lora}, BitFit \cite{zaken2021bitfit} and Diff-pruning \cite{guo2020diff}.

\keypoint{Prompt Tuning.} 
Inspired from the field of NLP, a rich line of recent works adapts VLPMs to downstream tasks by learning task-specific prompts in an end-to-end manner \cite{wang2022defo,chen2022prompt,zhu2022prompt}. 
Since only a handful of labeled examples are available during training, prompt tuning can be regarded as few-shot learning task \cite{zhang2023deta,zhang2022free}. 
In particular, CoOp \cite{zhou2022learning} performs task adaptation by  optimizing a set of prompt vectors at the language branch of CLIP. 
While simple and effective, CoOp tends to achieve poor generalization on new tasks after overfitting to the base (or target) task. 
To overcome this issue, CoCoOp \cite{zhou2022conditional} learns a lightweight meta-net to generate an input-conditional token for each input image.
By reducing the discrepancy between the hand-crafted prompt and the trainable prompt tokens, KgCoOp \cite{zhou2022conditional} significantly improves the generalization of the adapted models on new tasks. ProGrad \cite{zhu2022prompt} mitigates the overfitting issue by regularizing each tuning step that is not to conflict with the general knowledge of the hand-crafted prompt. 
Unlike the aforementioned methods that mainly focus on developing efficient textual prompts, a rich line of works also explores visual prompts for task adaptation \cite{jia2022vpt,huang2023diversity}.
By adding trainable prompts at both the language and text branches of CLIP, multi-model prompt tuning methods MaPLe\cite{khattak2023maple} and PromptSRC\cite{Khattak_2023_ICCV} yield remarkable performance on both the base task and new tasks.

\section{Conclusions}
In this work, we propose the DePT framework to tackle the Base-New Tradeoff (BNT) problem in prompt tuning.
First, we offer an insightful view to analyze the BNT problem, and reveal that the BNT stems from the channel bias issue. 
Second, we present the DePT framework to tackle the BNT problem, and DePT is orthogonal to existing prompt tuning methods.
Third, we apply DePT to a broad spectrum of baselines, and the results on 11 datasets demonstrate the strong flexibility and effectiveness of  DePT. 
We hope this work can bring some inspiration to related fields. 

\keypoint{Acknowledgements.} 
This work is supported by grants from the National Natural Science Foundation of China (Grant No. 62122018, No. 62020106008, No. U22A2097, No. U23A20315), Kuaishou Tech. and SongShan Laboratory YYJC012022019.

{
    \small
    \bibliographystyle{ieeenat_fullname}
    \bibliography{egbib}
}


\end{document}

%% file: preamble.tex
\usepackage[dvipsnames]{xcolor}
